\documentclass[11pt]{article}

\usepackage[final]{acl}

\usepackage{times}
\usepackage{latexsym}

\usepackage[T1]{fontenc}

\usepackage[utf8]{inputenc}

\usepackage{microtype}

\usepackage{inconsolata}

\usepackage{graphicx}
\usepackage{textcomp}
\usepackage{xcolor}
\usepackage{tcolorbox}
\usepackage{makecell}
\usepackage{amsmath,amssymb,amsfonts}
\usepackage{algorithmic}
\usepackage{multirow}
\usepackage{color}
\usepackage[table]{xcolor}
\usepackage{array}

\title{Exploring Lightweight Large Language Models for Court View Generation}

\author{Zhitian Hou$^1$,  Tianyong Hao$^2$, Nanli Zeng$^3$, Zhixiong Chao$^1$, Kun Zeng$^{1, \star}$ \\
        $^1$School of Computer Science and Engineering, Sun Yat-sen University \\
        $^2$School of Computer Science, South China Normal University\\
        $^3$China Mobile Internet Co., Ltd. \\
        \texttt{houzht@mail2.sysu.edu.cn;zengkun2@mail.sysu.edu.cn}
        }

\begin{document}
\maketitle
\begin{abstract}
Criminal Court View Generation (CVG) is a critical task in Legal Artificial Intelligence (Legal AI), involving the generation of court view based on case facts. In this work, we systematically explore the capabilities of lightweight (smaller than 2B) large language models (LLMs) in CVG and their impact on charge prediction. Our study addresses four key questions: (1) how does different architecture of LLMs affect the CVG quality and charge prediction. (2) how does LLMs size contribute to the performance, (3) how do lightweight LLMs compare with Deep Neural Networks (DNNs) in these tasks, and (4) how does predicting charge by court view generation first compare with predicting it directly. Additionally, we also develop CVGEvalKit, an evaluation framework including three public available datasets for CVG tasks, as well as predicting their charges. Comprehensive experiments are conducted on this framework, where models are trained on a mixed training set and evaluated on each dataset's test set. Experimental results provide new insights into the trade-offs between model architecture, model size, and the influence between different tasks, highlighting the potential of lightweight LLMs in judicial AI applications. The source code is anonymously available at \url{https://github.com/ZhitianHou/CVGEvalKit}
\end{abstract}

\begin{figure}[t]
  \centering
  \includegraphics[width=0.8\columnwidth]{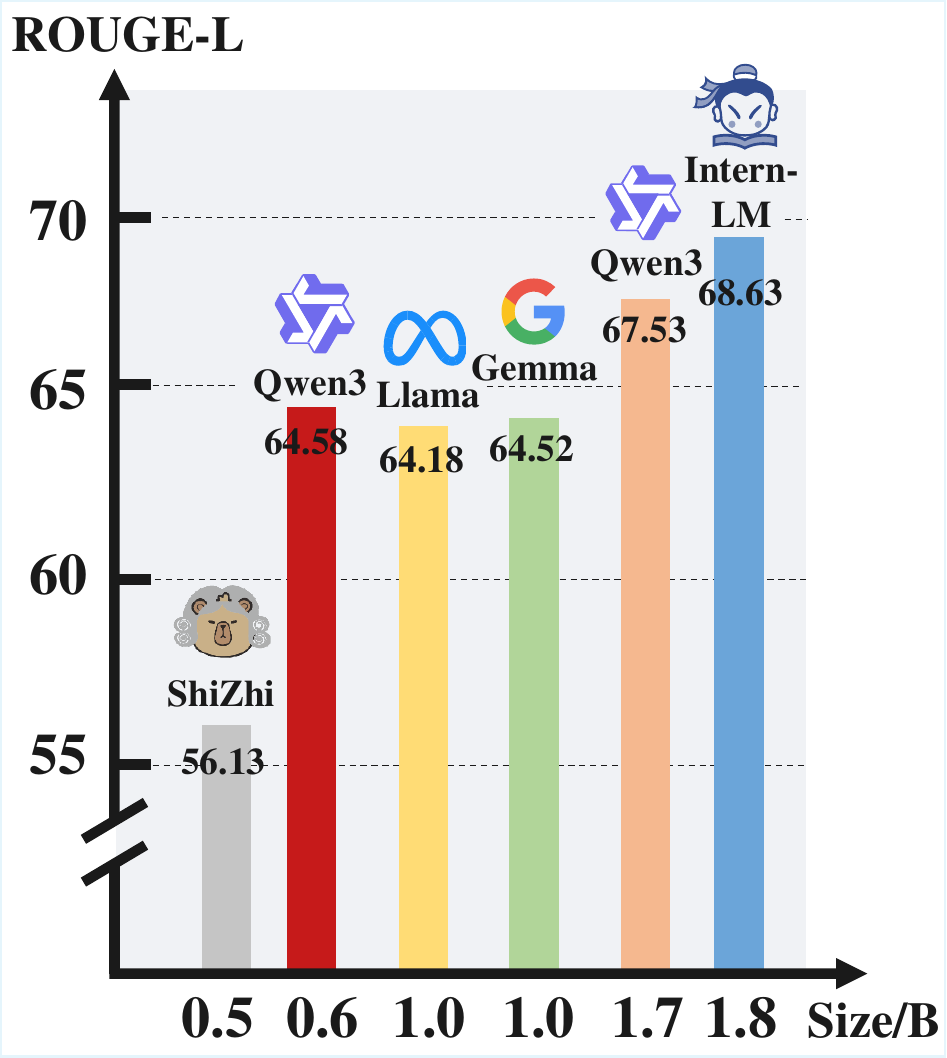}
  \caption {The average ROUGE-L of CVG of each test dataset of each model after fine-tuning.}
  \label{fig:average rouge-l}
\end{figure}

\section{Introduction}
In recent years, methods from the field of artificial intelligence have been increasingly applied to the legal domain \citep{do2024legal,gao2024legalformer,guo2024legal,hou2025large}. Criminal Court View Generation (CVG) is a crucial task in the field of legal artificial intelligence (Legal AI), aiming to automatically produce court view based on the fact description of a case \citep{ye2018interpretable}. In the legal system, a court view not only justifies the final judgment but also provides transparency and interpretability, ensuring that legal decisions are understandable and consistent. Traditionally, drafting court view requires significant human expertise and effort, which motivates the development of automated CVG systems. Beyond simply generating text, CVG plays a vital role in downstream applications, such as charge prediction, where the quality of court view directly affects the accuracy of subsequent decisions. Despite its importance, CVG remains a challenging problem due to the complexity of legal language, the need for logical consistency, and the necessity to incorporate statutory knowledge and factual details accurately.

Recent advances in natural language processing (NLP) have introduced Deep Neural Networks (DNNs) that can capture complex linguistic patterns and domain knowledge, and they have been applied to various legal tasks, including CVG \citep{yue2021circumstances, xu2024divide, yue2024event}. For example, Yue et al. \citep{yue2021circumstances} designed a circumstances-enhanced framework that separately generates adjudicating and sentencing reasoning. Xu et al. \citep{xu2024divide} proposed LeGen, which is a legal concept-guided framework for criminal CVG that incorporates key legal concepts such as recidivism, confession, and robbery. Yue et al. \citep{yue2024event} proposed Event Grounded Generation (EGG), which integrates fine-grained event information extracted from case facts into CVG. While DNNs have shown promise in language tasks, they often struggle to capture complex legal reasoning and long-range dependencies in case facts, which are crucial for generating accurate court views. 

On the other hand, Large Language Models (LLMs) exhibit stronger generative and reasoning capabilities, even in lightweight versions with moderate parameter scales, making them attractive candidates for CVG. As shown in Figure~\ref{fig:average rouge-l}, the open-source models can achieve better scores with bigger size after fine-tuning. However, systematic studies comparing successive LLM generations are scarce, and the effects of model architecture and size on both CVG quality and downstream applications, like charge prediction, remain underexplored. Additionally, the potential benefits of generating court reasoning before predicting charges versus direct charge prediction have not been thoroughly investigated, and comparative analyses between DNNs and LLMs on CVG tasks are largely missing. Addressing these gaps is essential to better understand the trade-offs in model design, scale, and task integration for judicial AI systems.

To address these gaps, this work conducts a systematic study on lightweight LLMs for CVG. Specifically, we investigate different models at diverse architectures and sizes to analyze their impacts on CVG task. We further examine how CVG affects downstream charge prediction by comparing two approaches: generating court reasoning before predicting charges versus direct charge prediction. Finally, we provide a comparative analysis of DNNs and LLMs on CVG tasks to highlight the advantages and limitations of lightweight LLMs in legal AI. Our contributions are fourfold: 
\begin{itemize}
    \item Analyzing the performance differences between different LLMs architecture.
    \item Studying the effect of LLMs size on CVG quality and charge prediction.
    \item Comparing DNNs and LLMs on CVG tasks, providing insights for the design of judicial AI systems.
    \item Evaluating the impact of court view generation on charge prediction accuracy.
\end{itemize}

\begin{figure*}[ht]
  \centering
  \includegraphics[width=0.9\textwidth]{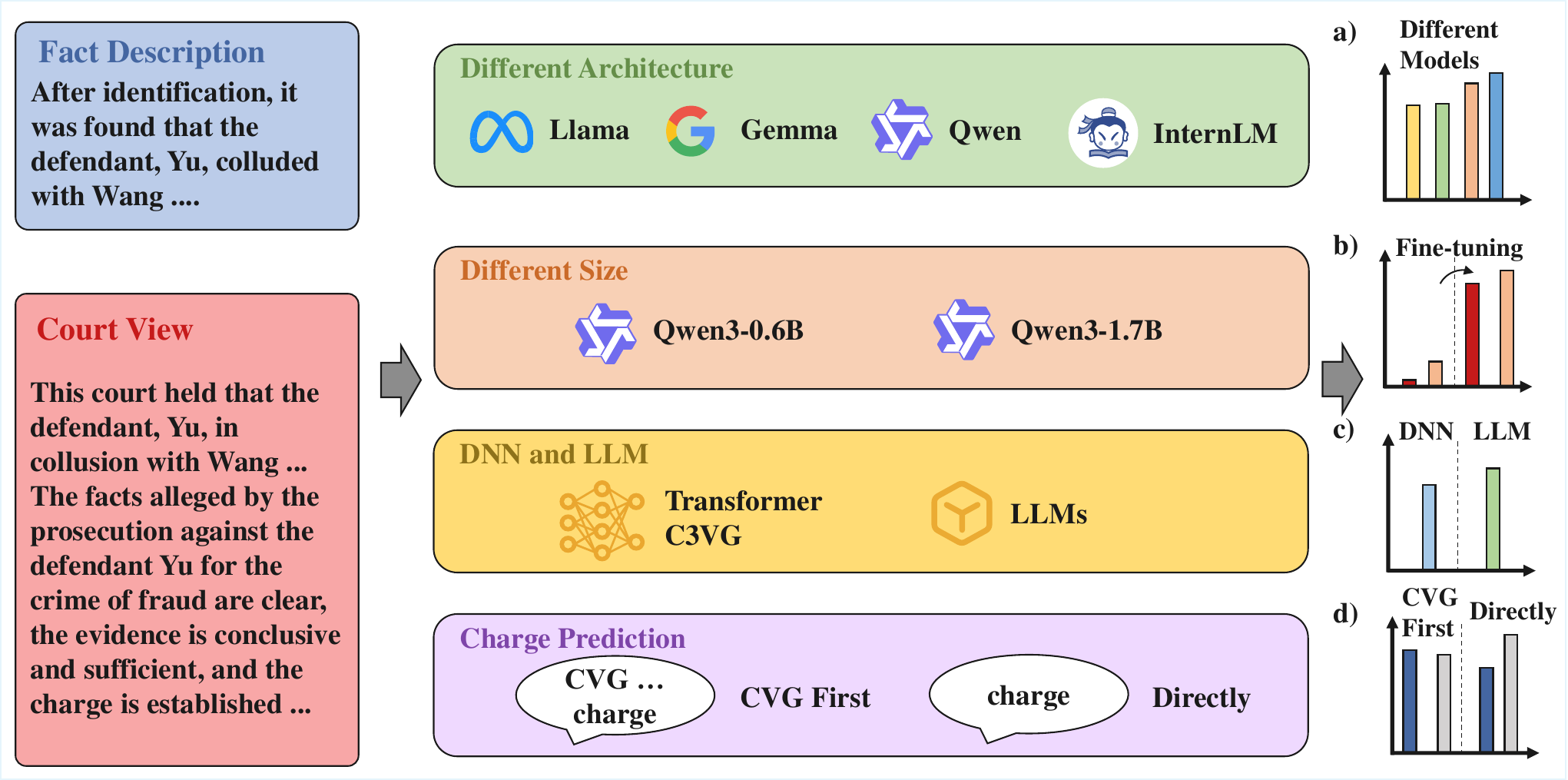}
  \caption {Overview of the research methodology and the four research questions investigated in this study.  
Subfigure (a) compares the performance across different model architectures 
(Llama-3.2-1B-Instruct, Gemma3-1B-IT, Qwen3-1.7B, and InternLM2.5-1.8B-Chat), addressing the effect of architecture. 
Subfigure (b) relects the impact of model size by contrasting Qwen3-0.6B and Qwen3-1.7B. 
Subfigure (c) compares the performance of DNN models and lightweight LLMs. 
Subfigure (d) compares two strategies for charge prediction, \textit{CVG First} and \textit{Directly}, where dark blue and gray bars represent the two settings respectively. 
All subfigures use ROUGE-L as the evaluation metric.}
  \label{fig:methodology}
\end{figure*}



\section{Related Work}
\subsection{Structured Approaches for Court View Generation}
Recent research efforts have increasingly focused on enhancing CVG through the explicit modeling of legal structures and the integration of domain-specific knowledge. Earlier works enhance generation quality by extracting detailed legal attributes, such as adjudicating and sentencing circumstances or domain-related legal concepts, from fact descriptions to enrich the relevance and factual grounding of generated court views \citep{yue2021circumstances, xu2024divide}. Other studies incorporate external legal knowledge, for instance by embedding law articles, charges, or claim information into the generation process \citep{ye2018interpretable, li2021court}, or by leveraging legal knowledge bases through prompt engineering and model guidance techniques \citep{li2024enhancing}. In addition, to improve interpretability and fairness, several methods adopt causality-driven reasoning frameworks, including counterfactual text generation \citep{wu2020biased, huang2023improving}, or employ modular generation pipelines with question–answer-based slot filling strategies \citep{huang2021generating}. Collectively, these efforts demonstrate that integrating structured legal knowledge and reasoning mechanisms can substantially enhance the coherence and reliability of CVG.

\subsection{Leveraging LLMs for Court View Generation.}
With the rapid development of LLMs, recent studies have begun exploring their potential for CVG and related legal reasoning tasks. Some research leverages general-purpose LLMs to identify intermediate legal structures from case facts \citep{yue2024event}, while others attempt to activate internal legal knowledge or incorporate external supervision signals to better adapt LLMs to legal text generation \citep{liu2024unleashing}. More recently, Hou et al. \citep{hou2025shizhichineselightweightlarge} introduces the first large language model, ShiZhi, specifically designed for CVG, demonstrating that even compact LLMs can produce coherent and legally grounded reasoning when trained on domain-specific data. However, though these works show that LLMs can be adapted for CVG, systematic analyses of lightweight LLMs and their performance on CVG tasks remain limited. In particular, there is a lack of comprehensive studies comparing different model architectures, model sizes, and their effects on both CVG quality and downstream applications.

\begin{table*}[htbp]
\centering
\small
\begin{tabular}{l|ccccc}
\hline
\textbf{Dataset} & \textbf{\# Train} & \textbf{\# Test}  & \makecell{\textbf{Mean Length}\\\textbf{of Fact}} & \makecell{\textbf{Mean Length}\\\textbf{of Court View}} & \textbf{Year Range} \\
\hline
C3VG  & 50,312 & 12,627  & 456.9  & 276.8  & 2012--2019 \\
\hline
LCVG  & 60,744 & 20,290 & 678.5  & 233.4  & 1998--2021 \\
\hline
CCVG  & 111,252 & 1,154 & 416.9  & 300.8  & 1985--2021 \\
\hline
\end{tabular}
\caption{Comparison of datasets used in our experiments. \# denotes the number.}
\label{tab:dataset_comparison}
\end{table*}

\section{Methodology}
\subsection{Overview}
As shown in Figure~\ref{fig:methodology}, this study systematically investigates lightweight LLMs for CVG and their influence on downstream charge prediction. We focus on the different open-source models at multiple architectures and sizes to analyze their performances. Our methodology consists of training models on a mixed CVG dataset, generating court views, and examining their impact on charge prediction under two paradigms.

\subsection{Model Architectures}
We consider multiple lightweight LLMs at different model architectures and sizes as representative models for CVG, including Qwen2.5-0.5B-Instruct \citep{qwen2.5}, Qwen3 \citep{yang2025qwen3}, Interm \citep{cai2024internlm2}, Llama-3.2-1B-Instruct, Gemma3-1B-IT \citep{team2025gemma}. DNN baselines adopt an encoder-decoder architecture, taking facts as input and generating court views. Compared with DNNs, LLMs have larger capacity and better long-range dependency modeling, which is crucial for legal reasoning. Comparing these models allows us to identify factors contributing to performance differences in CVG tasks.

\subsection{Data Preparation}
Our experiments involve three datasets: C3VG \citep{yue2021circumstances}, LCVG \citep{xu2024divide}, and CCVG \citep{hou2025shizhichineselightweightlarge}, each providing facts paired with corresponding court views. We construct a mixed training set by combining the training sets of these three datasets to assess model robustness and generalization. Each model is then evaluated on the test sets of three datasets separately. Standard preprocessing, including tokenization, normalization, and length truncation, is applied consistently across datasets.

\subsection{Training Procedure}
All models are fine-tuned using token-level cross-entropy loss to maximize the likelihood of the reference court views. For LLMs, we employ Low-Rank Adaptation (LoRA) \citep{hulora} for parameter-efficient fine-tuning, where only low-rank update matrices are trained while the original model weights remain frozen. The LoRA objective can be formulated as:

\begin{equation}
\begin{aligned}
\mathcal{L}_{\text{LoRA}} = - \sum_{t=1}^{T} 
\log P_\theta \big(y_t \mid y_{<t}, x; & W_0 + \Delta W \big), \\ & \Delta W = A B
\end{aligned}
\end{equation}

where $x$ denotes the input case facts and $y = (y_1, \dots, y_T)$ represents the reference response. $W_0 \in \mathbb{R}^{d \times k}$ is the original model weight matrix, while $\Delta W = AB$ is the low-rank update, with $A \in \mathbb{R}^{d \times r}$ and $B \in \mathbb{R}^{r \times k}$, where $r$ is the LoRA rank (we set $r=8$), $d$ is the input feature dimension, and $k$ is the output feature dimension of the weight matrix. Additional hyperparameters include the scaling factor $\alpha=32$ and dropout probability $0.05$.

\subsection{Task Definition and Research Questions}
The CVG task requires generating coherent and legally grounded court views from case facts. For charge prediction, we design two paradigms: a) Reasoning: generate court view first, then predict the charge based on the court view. b) No-Reasoning: predict the charge directly from case facts. For both paradigms, we adopt a structured prompt design to guide model outputs. The system prompt and query are shown in the following:

\begin{tcolorbox}[title=Prompt Design, colback=blue!10, colframe=black!80, boxrule=0.8pt]
\textbf{System Prompt:} You are a judge. Please generate the court views based on the case facts, and predict the corresponding charge according to the court views.\\
\textbf{Query:} Please output in the following format: \\
\textless view\textgreater \textit{court view}\textless /view\textgreater, \\ \textless charge\textgreater \textit{charge}\textless /charge\textgreater \\

Facts:\textbackslash n\{\textit{fact}\}\textbackslash nOutput:  

\end{tcolorbox}

For direct charge prediction, only the predicted charge needs to be output in the format \colorbox{gray!10}{\texttt{<charge>charge</charge>}}.

\begin{table*}[htbp]
\centering
\small
\begin{tabular}{lcccccccc}
\hline
\multirow{3}{*}{Models} & \multicolumn{6}{c}{Court View Generation} & \multicolumn{2}{c}{Charge} \\
    \cline{2-7}
    & \multicolumn{3}{c}{ROUGE} & \multicolumn{3}{c}{BLEU} & \multicolumn{2}{c}{Prediction} \\
    \cline{2-4} \cline{5-7} \cline{8-9}
    & R-1 & R-2 & R-L & B-1 & B-2 & B-N & Acc & MF1 \\
\hline
\rowcolor{gray!15}\multicolumn{9}{l}{\textit{DNN-based Models}} \\
AttS2S$^{*, \dagger}$ & 57.41 & 38.23 & 58.90 & 52.27 & 36.67 & 34.05 & -     & -     \\
Transformer$^{*, \dagger}$ & 61.05 & 40.67 & 58.45 & 51.86 & 39.40 & 35.78 & -     & -     \\
C3VG$^{*, \dagger}$ & 62.12 & 42.70 & 60.50 & 60.78 & 42.98 & 40.64 & 90.6  & 71.2  \\
\hline
\rowcolor{gray!15}\multicolumn{9}{l}{\textit{LLM-based Models}} \\ 
LeGen-PT(GLM3-6B)$^{*, \dagger}$ & 53.28 & 34.48 & 39.40 & 45.79 & 27.88 & 23.97 & -     & -     \\
LeGen-FT(GLM3-6B)$^{*, \dagger}$ & 74.32 & 64.88 & 65.60 & 57.03 & 53.78 & 51.43 & -     & -     \\
\hline
\rowcolor{gray!15}\multicolumn{9}{l}{\textit{Open-Source LLMs}} \\
Qwen3-0.6B & 3.36 & 0.35 & 2.61 & 0.14 & 0.12 & 0.09 & 59.87 & 74.90 \\
Qwen3-1.7B & 37.34 & 20.54 & 28.32 & 19.53 & 17.08 & 13.40 & 71.63 & 83.47 \\
Llama-3.2-1B-Instruct & 3.31 & 1.50 & 3.07 & 0.10 & 0.08 & 0.05 & 2.84 & 5.51 \\
Gemma3-1B-IT & 29.36 & 11.78 & 19.19 & 21.31 & 17.05 & 11.55 & 5.68 & 10.75 \\
InternLM2.5-1.8B-Chat & 3.01 & 0.17 & 2.35 & 0.00 & 0.00 & 0.00 & 0.03 & 0.06 \\
\rowcolor{gray!15}\multicolumn{9}{l}{\textit{Fine-tuning LLMs}} \\
ShiZhi$^{\dagger}$ & 62.01 & 43.21 & 53.28 & 59.03 & 52.37 & 44.18 & 88.85 & 94.10 \\ 
Qwen3-0.6B(FT) & 79.30 & 66.34 & 75.14 & 71.25 & 68.33 & 63.41 & 91.67 & 95.65 \\
Qwen3-1.7B(FT) & \underline{80.97} & \underline{68.85} & \underline{77.09} & \underline{73.20} & \underline{70.47} & \underline{65.88} & 93.28 & 96.52 \\
Llama-3.2-1B-Instruct(FT) & 78.72 & 64.66 & 74.09 & 69.99 & 66.87 & 61.58 & 93.36 & 96.56 \\
Gemma3-1B-IT(FT) & 79.03 & 64.99 & 74.37 & 70.56 & 67.43 & 62.11 & \textbf{93.86} & \textbf{96.83} \\
InternLM2.5-1.8B-Chat(FT) & \textbf{82.21} & \textbf{70.90} & \textbf{78.68} & \textbf{74.00} & \textbf{71.51} & \textbf{67.31} & \underline{93.53} & \underline{96.66} \\
\hline
\end{tabular}
\caption{Performance Comparison of Different Models on C3VG dataset. The best is \textbf{bolded}, and the second is \underline{underline}. Results marked with ''*'' are taken from ~\citep{yue2021circumstances, xu2024divide}, and those marked with ''$^\dagger$'' denote baseline models. FT means after fine-tuning.}
\label{tab:c3vg_results}
\end{table*}

Finally, our methodology supports addressing four research questions:
\begin{itemize}
    \item RQ1: How does different architecture of LLMs affect the CVG quality and charge prediction?
    \item RQ2: How does model size contribute to the performance?
    \item RQ3: How do lightweight LLMs compare with traditional Deep Neural Networks (DNNs) in these tasks?
    \item RQ4: How does predicting charge by court view generation first compare with predicting it directly?
\end{itemize}

\section{Experiments}
\subsection{Datasets}
We collected three CVG datasets to evaluate model performance. Table~\ref{tab:dataset_comparison} provides a comparison of these datasets, and the details of each are described below.

\noindent \textbf{C3VG} \citep{yue2021circumstances}. C3VG dataset is constructed from legal case documents on China Judgments Online\footnote{https://wenshu.court.gov.cn}, containing fact descriptions, charges, and rationales for both ADjudging Circumstance (ADC) and SEntencing Circumstance (SEC). We focus on single-charge cases by aligning fact sentences with SEC rationales. In total, C3VG includes 72,939 cases, split for extraction and generation stages.

\noindent \textbf{LCVG} \citep{xu2024divide}. LCVG is also derived from China Judgments Online and emphasizes legal concepts and their corresponding rationales. It includes the defendant's background, fact descriptions, legal concepts, and rationale for single-defendant cases.

\noindent \textbf{CCVG} \citep{hou2025shizhichineselightweightlarge}. CCVG is a large-scale court view generation dataset spanning 1985 to 2021. It contains rich fact descriptions and court views, filtered and preprocessed to ensure high quality.

\subsection{CVGEvalKit}
We develop CVGEvalKit, an extensible evaluation framework for systematically assessing models on both court view generation quality and charge prediction for multiple datasets. For evaluating court view generation, we employ ROUGE-1, ROUGE-2, and ROUGE-L and BLEU-1, BLEU-2, and BLEU-n to measure performance. To quantify charge prediction performance, we report accuracy (Acc) and F1 score. Specifically, for each test instance, the prediction is considered correct if the generated court view contains the correct charge label; otherwise, it is counted as incorrect. Acc and F1 are then computed over the whole single test set.

\subsection{Baselines}
\begin{itemize}
    \item \textbf{AttS2S}~\citep{bahdanau2014neural} is an attention-based encoder–decoder model for neural machine translation. It replaces a single fixed-length representation with a soft attention mechanism that lets the decoder focus on relevant source tokens during generation.
    \item \textbf{Transformer}~\citep{vaswani2017attention} is a self-attention-based encoder–decoder model that models global dependencies without recurrence or convolution, and is a widely used framework in several text generation tasks.
    \item \textbf{C3VG}~\citep{yue2021circumstances} selects sentences related to Adjudging Circumstance (ADC) and Sentencing Circumstance (SEC), and then generates court views for each type before merging them into the final output.
    \item \textbf{LeGen}~\citep{xu2024divide} is a model that decomposes court views into sub-views focused on specific legal concepts. It employs a concept reasoning module to generate targeted rationales and a verifier-generator pipeline to produce court views.
    \item \textbf{ShiZhi}~\citep{hou2025shizhichineselightweightlarge} is a large language model specifically designed for CVG, trained on a high-quality Chinese dataset (CCVG) of over 110K cases.
\end{itemize}

\begin{table*}[htbp]
\small
\centering
\begin{tabular}{lcccccccc}
\hline
\multirow{3}{*}{Models} & \multicolumn{6}{c}{Court View Generation} & \multicolumn{2}{c}{Charge} \\
    \cline{2-7}
    & \multicolumn{3}{c}{ROUGE} & \multicolumn{3}{c}{BLEU} & \multicolumn{2}{c}{Prediction} \\
    \cline{2-4} \cline{5-7} \cline{8-9}
    & R-1 & R-2 & R-L & B-1 & B-2 & B-N & Acc & MF1 \\
\hline
\rowcolor{gray!15}\multicolumn{9}{l}{\textit{DNN-based Models}} \\
AttS2S$^{*, \dagger}$ & 55.78 & 33.25 & 45.32 & 48.86 & 38.60 & 33.87 & -& -\\
Transformer$^{*, \dagger}$ & 55.45 & 30.20 & 42.45 & 49.05 & 34.43 & 30.78 & -& -\\
C3VG$^{*, \dagger}$ & 57.20 & 37.45 & 45.50 & 50.12 & 40.52 & 37.64 & -& -\\
\hline
\rowcolor{gray!15}\multicolumn{9}{l}{\textit{LLM-based Models}} \\ 
LeGen-PT (GLM3-6B)$^{*, \dagger}$ & 41.54 & 22.93 & 28.33 & 34.88 & 13.47 & 10.97 & -& -\\
LeGen-FT (GLM3-6B)$^{*, \dagger}$ & \textbf{77.76} & \textbf{67.80} & 63.44 & 60.78 & 54.98 & 51.88 & -& -\\
\hline
\rowcolor{gray!15}\multicolumn{9}{l}{\textit{Open-Source LLMs}} \\
Qwen3-0.6B & 5.58 & 1.05 & 4.34 & 1.22 & 0.99 & 0.69 & 65.17 & 78.92 \\
Qwen3-1.7B & 20.26 & 9.57 & 15.47 & 16.24 & 13.54 & 9.98 & 54.14 & 70.25 \\
Llama-3.2-1B-Instruct & 3.13 & 1.13 & 2.85 & 0.23 & 0.16 & 0.09 & 0.08 & 0.16 \\
Gemma3-1B-IT & 23.83 & 8.43 & 16.83 & 14.47 & 11.58 & 7.43 & 2.34 & 4.58 \\
InternLM2.5-1.8B-Chat & 3.54 & 0.04 & 2.82 & 0.00 & 0.00 & 0.00 & 0.08 & 0.17 \\
\rowcolor{gray!15}\multicolumn{9}{l}{\textit{Fine-tuning LLMs}} \\ 
ShiZhi$^{\dagger}$ & 63.68 & 45.39 & 55.49 & 61.83 & 55.60 & 47.31 & 92.83 & 96.28 \\
Qwen3-0.6B(FT) & 67.47 & 51.64 & 61.19 & 60.79 & 56.87 & 50.54 & 89.93 & 94.70 \\
Qwen3-1.7B(FT) & 70.17 & 54.74 & \underline{64.23} & \textbf{62.75} & \textbf{58.99} & \underline{52.92} & \underline{92.86} & \underline{96.30} \\
Llama-3.2-1B-Instruct(FT) & 67.69 & 50.59 & 60.82 & 60.27 & 56.23 & 49.57 & 92.14 & 95.91 \\
Gemma3-1B-IT(FT) & 67.71 & 50.72 & 60.98 & 60.86 & 56.72 & 49.97 & 92.40 & 96.05 \\
InternLM2.5-1.8B-Chat(FT) & \underline{71.55} & \underline{57.51} & \textbf{66.61} & \underline{61.12} & \underline{58.08} & \textbf{53.04} & \textbf{93.50} & \textbf{96.64} \\
\hline
\end{tabular}
\caption{Performance Comparison of Different Models on LCVG dataset. The best is \textbf{bolded}, and the second is \underline{underline}. Results marked with ''*'' are taken from ~\citep{yue2021circumstances, xu2024divide}, and those marked with ''$^\dagger$'' denote baseline models. FT means after fine-tuning.}
\label{tab:lcvg_results}
\end{table*}

\begin{figure}[t]
  \centering
  \includegraphics[width=\columnwidth]{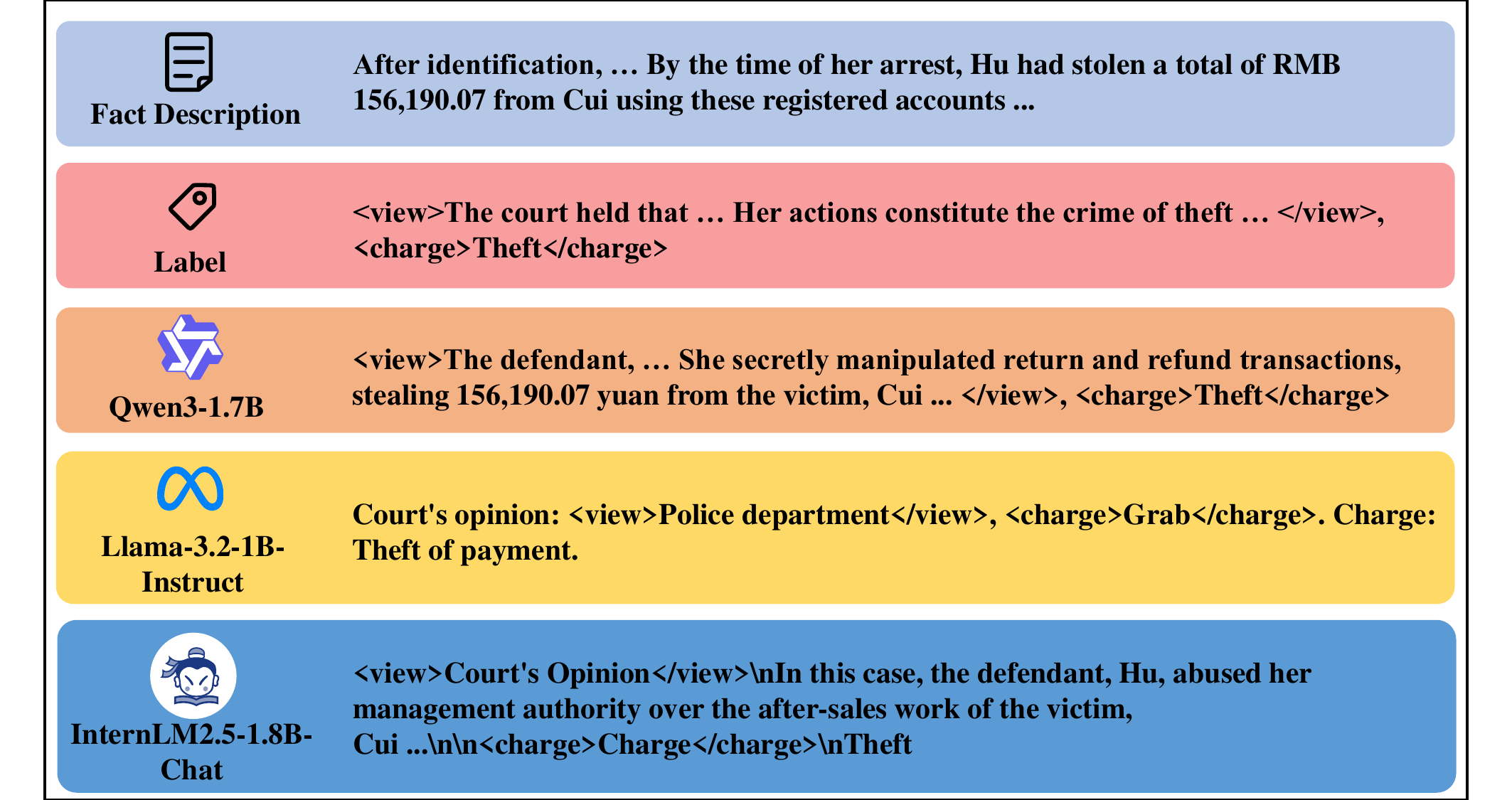}
  \caption {The examples of Qwen3-1.7B, Llama-3.2-1B-Instruct and InternLM2.5-1.8B-Chat in zero-shot setting.}
  \label{fig:rq1_case}
\end{figure}

\subsection{RQ1: How does different architecture of LLMs affect the CVG quality and charge prediction?}
The results shown in Table~\ref{tab:c3vg_results}, Table \ref{tab:lcvg_results}, and Table \ref{tab:ccvg_results} clearly show that model architecture plays a crucial role in both court view generation and charge prediction. Among the zero-shot open-source LLMs, Qwen3-1.7B generally outperform Llama-3.2-1B-Instruct and InternLM2.5-1.8B-Chat in ROUGE and BLEU metrics, achieving 22.03 and 11.49 on average, respectively. This is likely because the Qwen3 architecture is specifically designed for instruction-following and efficient reasoning, enabling better understanding of legal facts and context. The examples of these models are shown in Figure~\ref{fig:rq1_case}. Compared to responses from other models, Qwen3-1.7B perfectly follows the template specified in the prompt. Interestingly, Gemma3-1B-IT also obtaining 18.66 and 9.62 across three datasets in CVG task despite its relatively small size and the fact that much of its training data is not Chinese. This may benefit from its instruction-tuning and exposure to reasoning-intensive data (e.g., mathematics), which improves text organization and logical consistency. However, its responses to charges are often semantically close to the gold labels but not expressed using the correct legal terminology. This may be because the model lacks sufficient exposure to Chinese legal texts during pre-training, resulting in gaps in its knowledge of domain-specific terminology.

\begin{table*}[htbp]
\centering
\small
\begin{tabular}{lcccccccc}
\hline
\multirow{3}{*}{Models}  & \multicolumn{6}{c}{Court View Generation} & \multicolumn{2}{c}{Charge} \\
    \cline{2-7}
    & \multicolumn{3}{c}{ROUGE} & \multicolumn{3}{c}{BLEU} & \multicolumn{2}{c}{Prediction} \\
    \cline{2-4} \cline{5-7} \cline{8-9}
    & R-1 & R-2 & R-L & B-1 & B-2 & B-N & Acc & MF1 \\
\hline
\rowcolor{gray!15}\multicolumn{9}{l}{\textit{Zero-shot LLMs}} \\ 
Qwen3-0.6B & 4.60 & 0.79 & 3.76 & 0.58 & 0.48 & 0.34 & 64.64 & 78.53 \\
Qwen3-1.7B & 30.24 & 15.04 & 22.31 & 17.11 & 14.73 & 11.10 & 63.26 & 77.49 \\
Llama-3.2-1B-Instruct & 1.64 & 0.68 & 1.50 & 0.02 & 0.01 & 0.01 & 0.17 & 0.35 \\
Gemma3-1B-IT & 28.79 & 11.63 & 19.96 & 17.26 & 14.21 & 9.88 & 5.11 & 9.73 \\
InternLM2.5-1.8B-Chat & 3.00 & 0.06 & 2.63 & 0.00 & 0.00 & 0.00 & 0.09 & 0.17 \\
\rowcolor{gray!15}\multicolumn{9}{l}{\textit{Fine-tuning LLMs}} \\ ShiZhi${^{\dagger}}$ & \textbf{70.00} & 51.07 & 59.61 & \textbf{67.85} & \textbf{62.92} & 54.76 & \underline{86.48} & 92.75 \\
Qwen3-0.6B(FT) & 65.66 & 49.23 & 57.41 & 63.21 & 58.72 & 51.51 & 80.07 & 88.93 \\
Qwen3-1.7B(FT) & \underline{69.58} & \underline{52.92} & \textbf{61.27} & 66.68 & 62.10 & \underline{54.83} & \textbf{86.74} & \underline{92.90} \\
Llama-3.2-1B-Instruct(FT) & 66.84 & 48.85 & 57.62 & 64.14 & 59.37 & 51.59 & 84.23 & 91.44 \\
Gemma3-1B-IT(FT) & 67.40 & 49.53 & 58.22 & 65.14 & 60.28 & 52.42 & 84.32 & \textbf{94.49} \\
InternLM2.5-1.8B-Chat(FT) & 69.04 & \textbf{53.05} & \underline{60.60} & \underline{66.70} & \underline{62.25} & \textbf{55.22} & 84.23 & 91.44 \\
\hline
\end{tabular}
\caption{Performance Comparison of Different Models on CCVG dataset. The best is \textbf{bolded}, and the second is \underline{underline}. Results marked with ''$^\dagger$'' denote baseline models. FT means after fine-tuning.}
\label{tab:ccvg_results}
\end{table*}

\begin{figure}[t]
  \centering
  \includegraphics[width=0.7\columnwidth]{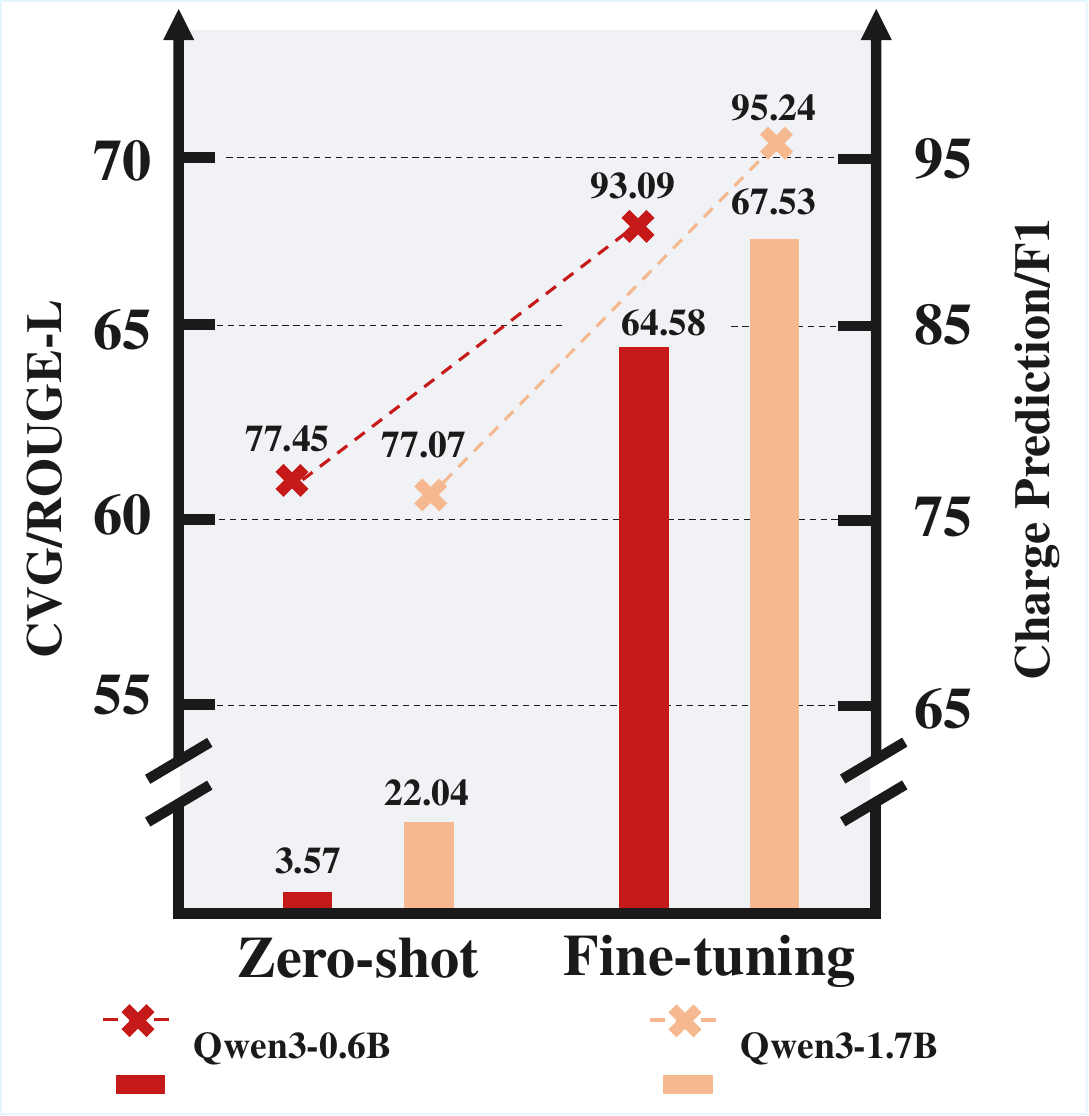}
  \caption {The performance of Qwen3-0.6B and Qwen3-1.7B. The dashed line denotes the model’s F1 score on charge prediction, while the bars indicate ROUGE-L for the CVG task, computed as the average across the three datasets.}
  \label{fig:rq2_plot}
\end{figure}

Fine-tuning further enhances performance across all architectures, indicating that both the inherent model structure and the ability to adapt to domain-specific instructions are key factors. Notably, InternLM2.5-1.8B-Chat achieves the best performance after fine-tuning with 68.63 on ROUGE-L and 58.52 on BLEU-N, but performs poorly in zero-shot settings, which is likely affected by its reliance on strict prompt formats. 

Overall, models with architectures optimized for instruction following, multi-step reasoning, and context-aware encoding produce more accurate and fluent court views and higher charge prediction scores. 

\begin{figure}[t]
  \centering
  \includegraphics[width=\columnwidth]{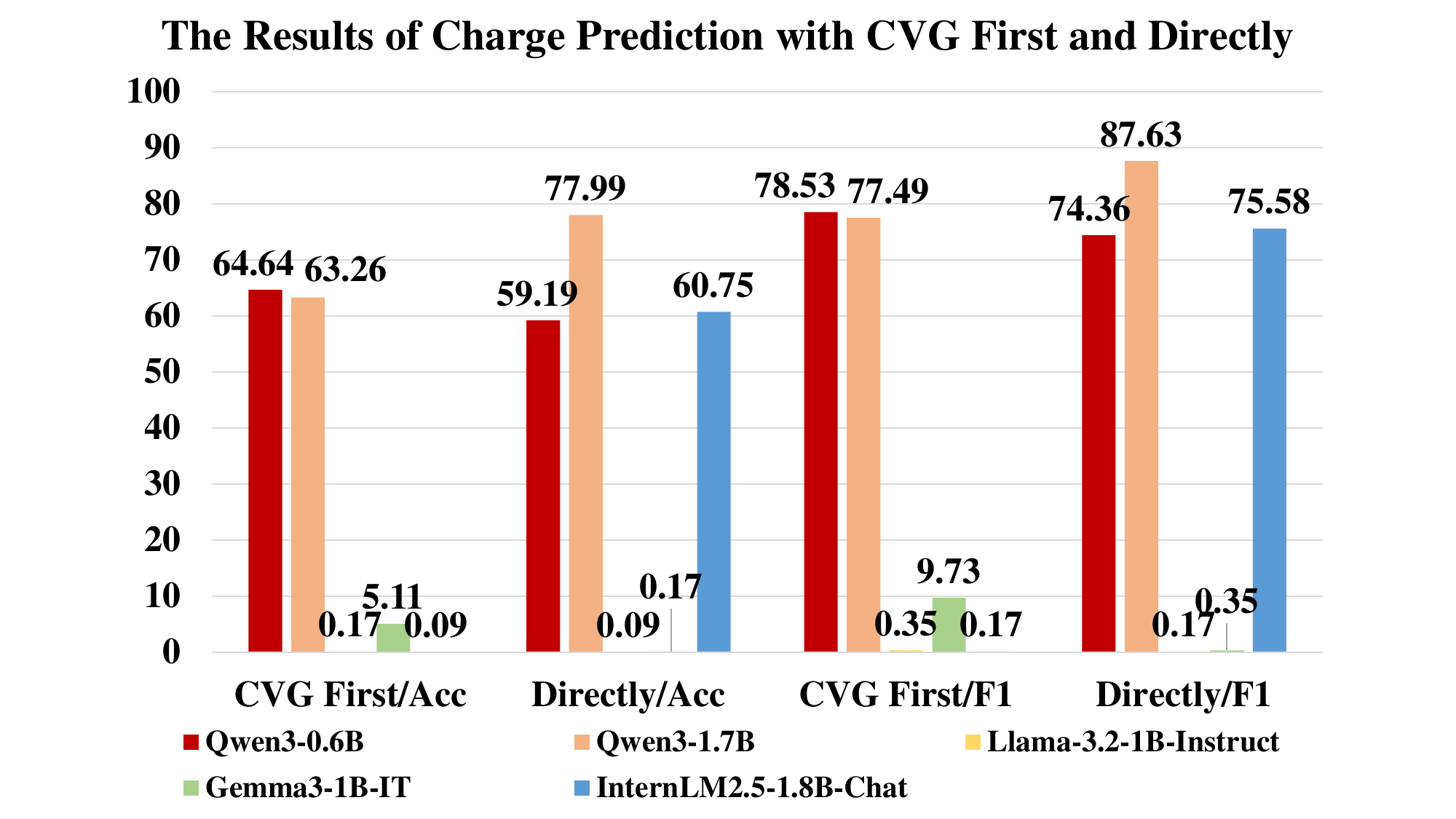}
  \caption {The results of charge prediction with CVG first and directly.}
  \label{fig:cvgfirst_results}
\end{figure}

\subsection{RQ2: How does model size contribute to the performance?}
For the Qwen3 family, where the architecture is kept constant, model size shows a clear effect on performance as shown in Figure~\ref{fig:rq2_plot}. For CVG task, In the zero-shot setting, the 1.7B model outperforms the 0.6B model by roughly 7$\times$ of ROUGE score, indicating that parameter capacity has a substantial impact on both court view generation. After fine-tuning, the gap between 0.6B and 1.7B becomes much smaller, suggesting that domain-specific adaptation can compensate for part of the advantage brought by larger model size. Disregarding architectural differences and focusing solely on model size, the results in Figure~\ref{fig:average rouge-l} show that performance after fine-tuning is roughly positively correlated with model size. However, since most open-source models in our experiments only provide a single size below 2B parameters, we are not able to draw broader, architecture-agnostic conclusions about model scaling.

\begin{figure}[t]
  \centering
  \includegraphics[width=0.9\columnwidth]{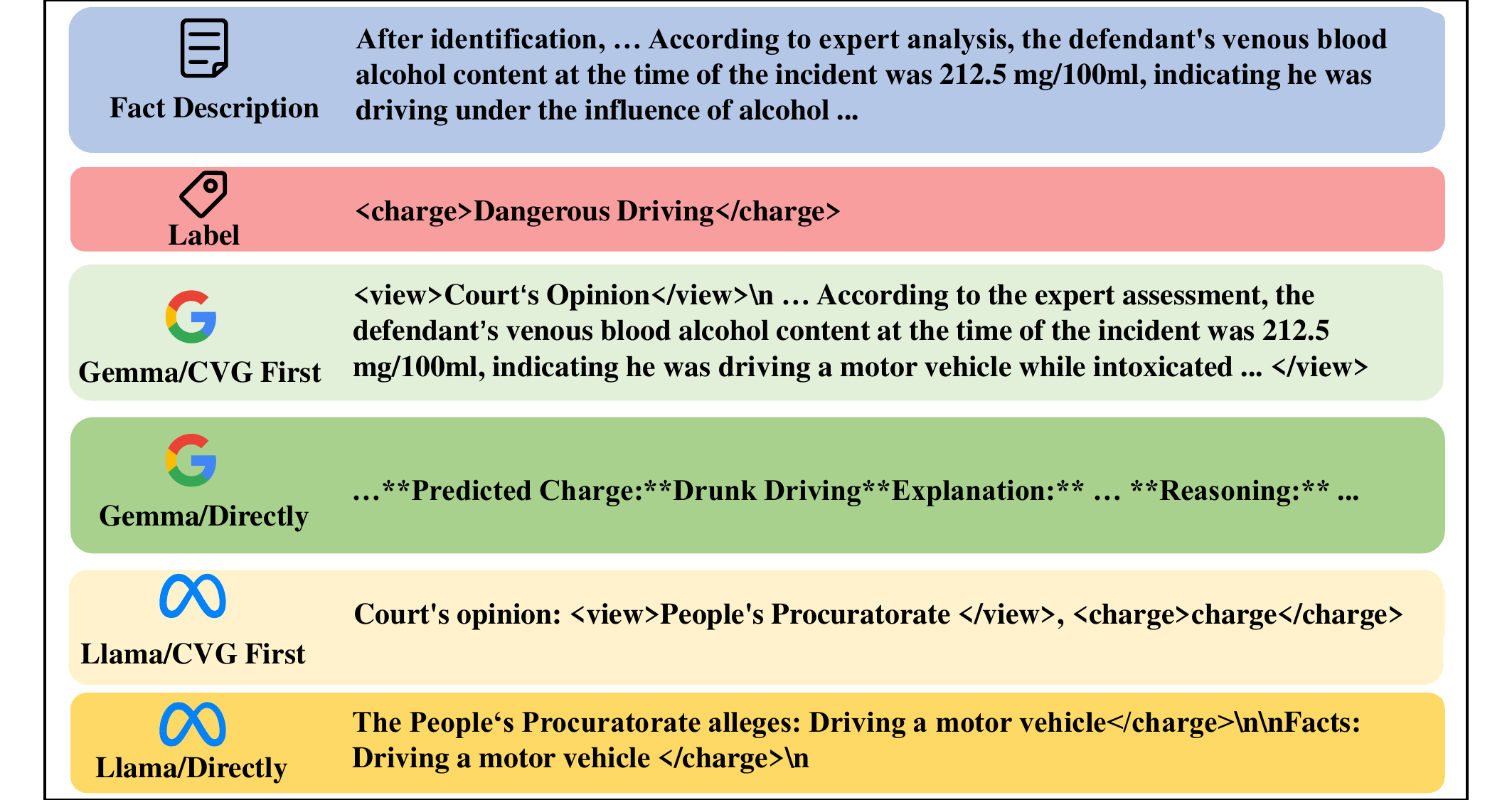}
  \caption {The example of Gemma3-1B-IT and Llama-3.2-1B-Instruct in charge prediction with CVG first and directly.}
  \label{fig:gemma_llama_cp_case}
\end{figure}

For charge prediction task, the results in Figure~\ref{fig:rq2_plot} show that the performance gap between the two model sizes is small under the zero-shot setting. This indicates that models in the Qwen3 family exhibit relatively strong charge prediction capability regardless of their sizes, possibly due to exposure to legal judgments during training. However, the gap becomes wider with the relative difference increasing from 0.5\% to 2\% after fine-tuning, suggesting that larger models benefit more from fine-tuning and can better internalize knowledge from gold-standard answers. 

Across different model architectures, performance on the charge prediction task does not exhibit a clear correlation with model size, suggesting that training data and architectural choices play a more crucial role than size alone. Under the zero-shot setting, all models except Qwen3 show almost collapsed performance on charge prediction, indicating little connection to model size. After fine-tuning, however, all models converge to very similar performance levels, with only a slight positive correlation with size. Notably, our experimental results reveal that below 1B parameters, increasing model size does not yield performance gains, whereas beyond the 1B threshold, performance improves consistently with scale. This suggests that the Scaling Law has certain boundaries and does not hold uniformly across all model sizes.

\begin{figure}[t]
  \centering
  \includegraphics[width=0.9\columnwidth]{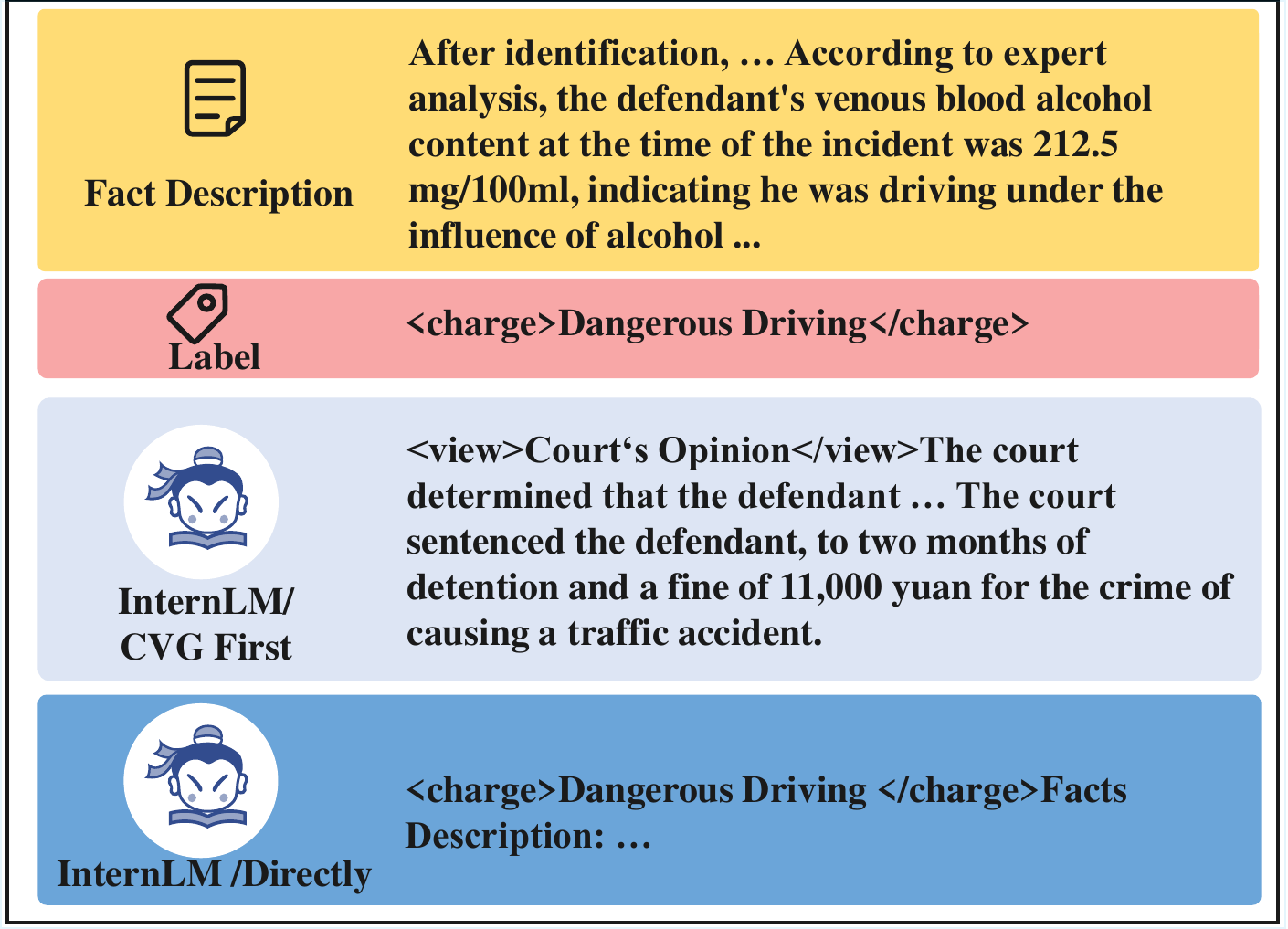}
  \caption {The example of InternLM2.5-1.8B-Chat in charge prediction with CVG first and directly.}
  \label{fig:internlm_cp_case}
\end{figure}

\subsection{RQ3: How do lightweight LLMs compare with Deep Neural Networks (DNNs) in these tasks?}
As shown in Table~\ref{tab:c3vg_results} and Table~\ref{tab:lcvg_results}, we observe that fine-tuned lightweight LLMs consistently outperform traditional DNN-based models on both court view generation and charge prediction. For example, InternLM2.5-1.8B-Chat(FT) achieves 78.68 on ROUGE-L and 67.31 on BLEU-N, which represents gains of approximately 32\% and 66\% over the strongest DNN baseline (C3VG). On charge prediction, its Acc and Macro-F1 further exceed C3VG by about 3\% and 36\%, respectively. In addition, the results in Table~\ref{tab:ccvg_results} indicates that models such as LeGen-FT and ShiZhi show strong performance on the datasets they are specifically fine-tuned on, but their performance drops substantially on other datasets, indicating limited cross-dataset generalization. In contrast, models fine-tuned on multiple datasets maintain more stable improvements, suggesting that multi-dataset fine-tuning provides better robustness and transferability than single-dataset specialization.

\subsection{RQ4: How does predicting charge by court view generation first compare with predicting it directly?}
As shown in Figure~\ref{fig:cvgfirst_results}, only the Qwen3 series demonstrates stable and accurate charge prediction under the zero-shot setting, regardless of whether the model is asked to generate a court view first or to directly output the charge. In contrast, as illustrated in Figure~\ref{fig:gemma_llama_cp_case}, both Llama and Gemma almost fail on charge prediction in zero-shot mode. They struggle to follow instructions, frequently produce outputs in incorrect formats, and show very limited accuracy. Overall, the results show that models with 1B parameters or below tend to benefit from the “CVG first” strategy, with Qwen3-0.6B and Gemma3-1B-IT showing the clearest improvements. However, for larger lightweight models such as Qwen3-1.7B and InternLM2.5-1.8B-Chat, generating court views before predicting the charge leads to performance degradation. In particular, InternLM2.5-1.8B-Chat appears to be strongly disturbed when required to perform the two-step instruction, suggesting that its ability to handle complex multi-stage prompts is still limited, and it performs better when focusing on a single, well-defined prediction task, as also reflected in Figure~\ref{fig:internlm_cp_case}.

\section{Conclusion}
In this paper, we investigated the performance of lightweight ($<$2B) LLMs on the Court View Generation (CVG) task, as well as the impact of charge prediction. Our study leads to several key findings. 1) Training paradigms and architectures substantially affect performance on both CVG and charge prediction. 2) Model size plays an important role in CVG, whereas its impact on charge prediction is comparatively limited. 3) After fine-tuning, LLMs can significantly outperform traditional DNN-based models, highlighting the potential of LLMs for court view generation. 4) For models with $\leq$1B parameters, generating court views prior to charge prediction is beneficial, while for larger models ($>$1B), this strategy tends to hurt charge prediction performance. In addition, we introduce CVGEvalKit, an extensible evaluation framework designed to systematically assess model performance on both CVG and charge prediction, facilitating future research in legal AI.

\section*{Limitations} \label{sec: limitations}
This work has two main limitations. First, due to structural differences in legal documents, our study is conducted only on Chinese datasets, and results may not directly generalize to other languages or legal systems. Second, our conclusions are based on lightweight LLMs ($\leq$2B) and may not hold for larger models. In future work, we plan to extend our experiments to multilingual and cross-jurisdictional datasets, as well as explore the behaviour of larger LLMs on court view generation and charge prediction tasks.

\section*{Ethical considerations}
This paper explores the use of LLMs for Court View Generation and charge prediction. Since the tasks involve sensitive legal information, the models should be used solely for research or supportive purposes, not for actual judicial decisions. Care must be taken to ensure privacy, avoid misuse, and prevent the outputs from being interpreted as authoritative legal advice.

\bibliography{custom}

\end{document}